\RequirePackage{snapshot}
 \documentclass[tablecaption=bottom,wcp]{jmlr} 

\usepackage{booktabs}
\usepackage{float}
\usepackage{subfiles}
\usepackage{braket}
\usepackage{xcolor}
\usepackage{cleveref}
\usepackage[load-configurations=version-1]{siunitx} 

\theorembodyfont{\upshape}
\theoremheaderfont{\scshape}
\theorempostheader{:}
\theoremsep{\newline}

\newcommand{\X}{\mathcal{X}}
\newcommand{\Y}{\mathcal{Y}}
\newcommand{\K}{\mathcal{K}}
\newcommand{\N}{\mathcal{N}}
\newcommand{\KL}{D_\textrm{KL}}
\newcommand{\ReLU}{\textsc{ReLU}}
\newcommand{\Erf}{\textsc{Erf}}
\newcommand{\kl}[2]{\KL\left[ #1 \Vert #2 \right]}
\DeclareMathOperator{\Tr}{Tr}

\jmlrproceedings{AABI 2019}{2nd Symposium on Advances in Approximate Bayesian Inference, 2019}

\title[Information in Infinite Nets]{Information in Infinite Ensembles of Infinitely-Wide Neural Networks}

\author{
    \Name{Ravid Shwartz-Ziv}\nametag{\thanks{Work done while an intern at Google Research.}} \Email{ravid.ziv@mail.huji.ac.il} \\
    \addr Edmond and Lilly Safra Center for Brain Sciences \\
    \addr The Hebrew University of Jerusalem \\
    \AND
    \Name{Alexander A. Alemi} \Email{alemi@google.com}\\
    \addr Google Research \\
}

\begin{document}

\maketitle

\begin{abstract}
    In this preliminary work, we study the
    generalization properties
    of infinite ensembles of infinitely-wide neural networks.
    Amazingly, this model family admits tractable calculations 
    for many information-theoretic quantities.
    We report analytical and empirical investigations 
    in the search for signals that correlate with generalization.
\end{abstract}


\section{Introduction}

A major area of research is to understand 
deep neural networks'
remarkable ability to generalize to unseen examples.
One promising research direction is to view deep neural networks through the lens of information 
theory~\citep{tishbydeep}. Abstractly, deep connections exist between the information
a learning algorithm extracts and its generalization
capabilities~\citep{littlebits, bayesianbounds}.  Inspired by these general results,
recent papers have attempted to measure information-theoretic quantities in ordinary
deterministic neural networks~\citep{blackbox,emergence,whereinfo}.  

Both practical and theoretical problems arise 
in the deterministic case~\citep{hownot, saxe, brendan}.
These difficulties stem from the fact that 
mutual information (MI) is reparameterization independent~\citep{coverthomas}.\footnote{
This implies
that if we send a random variable through an invertible function, its MI
with respect to any other variable remains unchanged.} 
One workaround is to make a network explicitly stochastic, either in its activations~\citep{vib} or
its weights~\citep{emergence}.
Here we take an alternative approach, harnessing 
the stochasticity in our 
choice of initial parameters.
That is, we consider
an \emph{ensemble} of neural networks, all trained with the same training procedure and data.
This will generate an ensemble of predictions.
Characterizing the generalization properties of the ensemble 
should characterize the generalization of 
individual draws from this ensemble.

%

Infinitely-wide neural networks behave as if
they are linear in their parameters~\citep{widelinear}.
Their evolution is fully 
described by the 
\emph{neural tangent kernel} (NTK). The NTK
is constant in time and can be tractably
computed~\citep{neuraltangents}.
For our purposes, it
can be considered to be a function of the network's
architecture, e.g. the number and the structure of layers, nonlinearity, initial parameters' distributions, etc.

All told, the output of an infinite ensemble of
infinitely-wide neural networks initialized with Gaussian weights and biases
and trained with gradient flow to minimize a square loss is
simply a conditional Gaussian distribution:
\begin{equation}
    p(z|x) \sim \N(\mu(x,\tau), \Sigma(x,\tau)) ,
    \label{eqn:rep}
\end{equation}
where $z$ is the output of the network and $x$ is its input.
The mean $\mu(x,\tau)$ and covariance $\Sigma(x, \tau)$ functions
can be computed~\citep{neuraltangents}.
For more background on the NTK and NNGP as well
as full forms of $\mu$ and $\Sigma$, see~\cref{sec:ntk}.

This simple form allows us to bound
several interesting information-theoretic quantities including:
the MI between the representation and the targets ($I(Z;Y)$, \cref{sec:izy}),
the MI between the representation and the inputs after training ($I(Z;X|D)$, \cref{sec:izx}),
and the MI between the representations and the training set, conditioned on the input ($I(Z;D|X)$, \cref{sec:izd}),
We are also able to compute in closed form:
the Fisher information metric~(\cref{sec:fisher}),
the distance the parameters move~(\cref{sec:dist}),
and the MI between the parameters and the data ($I(\Theta;D)$, \cref{sec:itd}).
Because infinitely-wide neural networks are linear in their parameters, 
their information geometry in parameter space is very simple.
The Fisher information metric is constant and flat, so the trace of the Fisher does not
evolve as in~\citet{whereinfo}. While the Euclidean distance the parameters move is small~\citep{widelinear}, the distance they move according to the Fisher metric is finite.
Finally, the MI between the data and the parameters tends to infinity, rendering
PAC Bayes style bounds on generalization vacuous~\citep{emergence,bayesianbounds,littlebits}.

\section{Experiments}



For jointly Gaussian data (inputs $X$ and targets $Y$), 
the Gaussian Information Bottleneck~\citep{gaussib} gives
an exact characterization of the optimal tradeoff between $I(Z;X)$ and $I(Z;Y)$,
where $Z$ is a stochastic representation, $p(z|x)$, of the input.
Below we fit infinite ensembles of infinitely-wide neural networks to jointly Gaussian
data and measure estimates of these mutual informations.
This allows us to assess how close to optimal
these networks perform.

The Gaussian dataset we created (for a details, see~\cref{sec:gaussian})
has $|X|=30$ and $|Y|=1$. We trained 
a three-layer FC network 
with both \ReLU\ and \Erf\ activation functions. 

\begin{figure}[htb]
  \centering
  \subfigure[ReLU]{\includegraphics[width=0.45\textwidth]{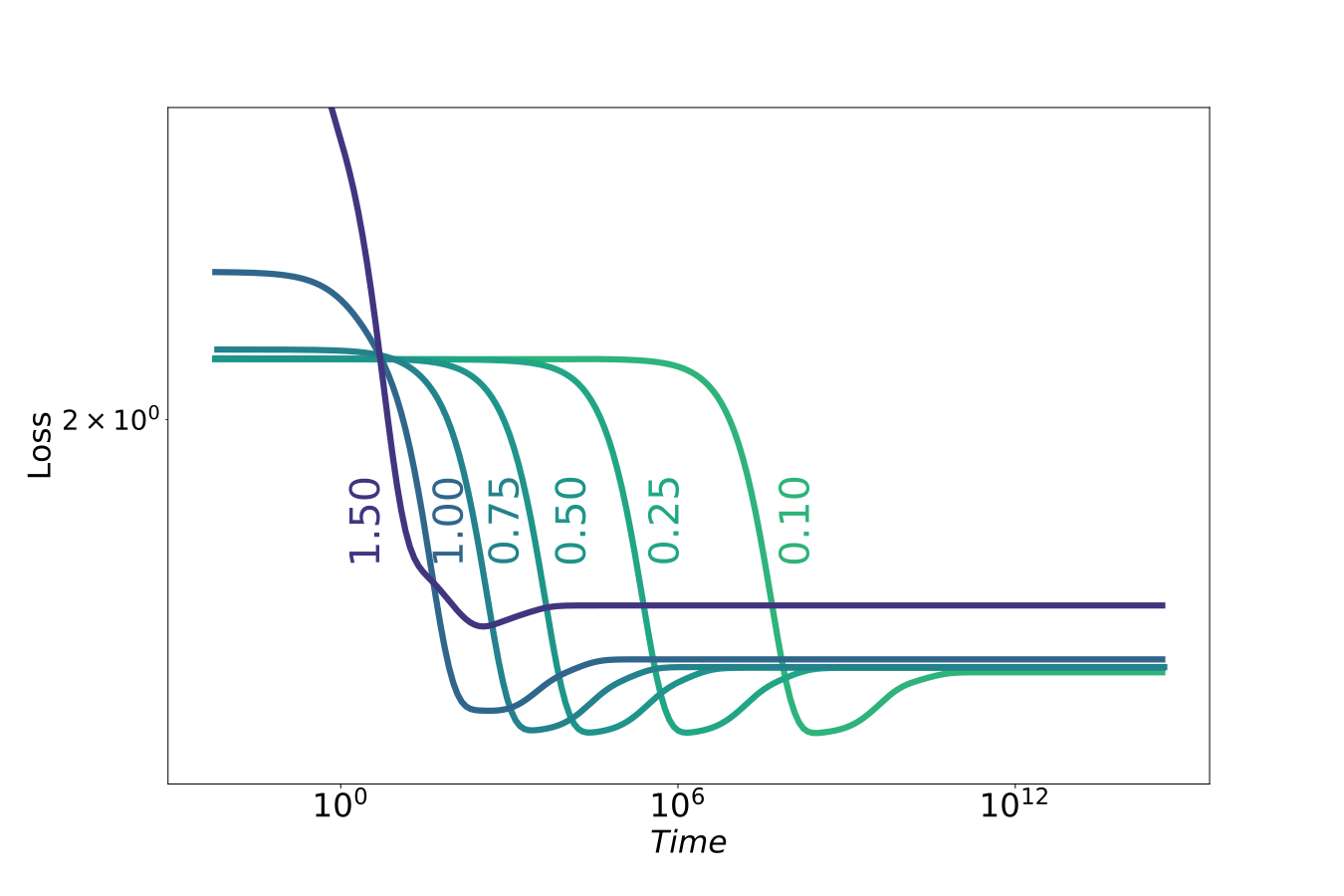} \label{fig:loss_vs_time_gauss_relu}}
  \subfigure[Erf]{\includegraphics[width=0.45\textwidth]{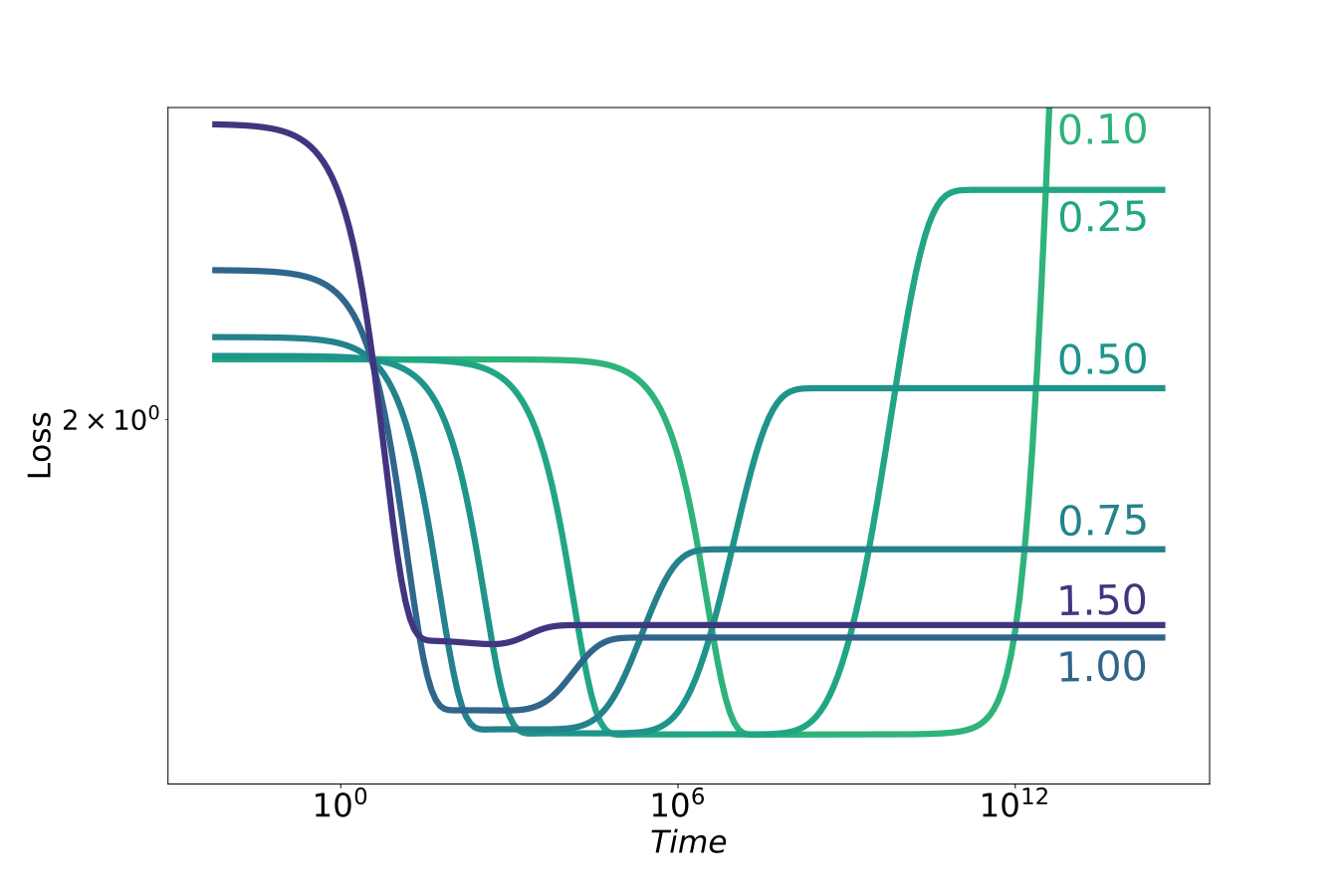}\label{ig:loss_vs_time_gauss_erf}}
        \caption{Loss as function of time for different initial weights' variances on the Gaussian dataset.}
         \label{fig:loss_vs_time_gauss}
\end{figure}

\Cref{fig:loss_vs_time_gauss} shows
the test set loss as a function of time for different choices of initial
weight variance ($\sigma_w^2$).  
For both the \ReLU\ and \Erf\ networks, at the highest $\sigma_w$ shown (darkest purple),
the networks \emph{underfit}. 
For lower initial weight variances, they all show signs of \emph{overfitting} in the sense that
the networks would benefit from early stopping.  This overfitting is worse
for the \Erf\ non-linearity 
where we see a divergence in the final test set loss as $\sigma_w$ decreases.
For all of these networks the training loss goes to zero.

\begin{figure}[htb]
  \centering
  \subfigure[\ReLU]{\includegraphics[width=0.45\textwidth]{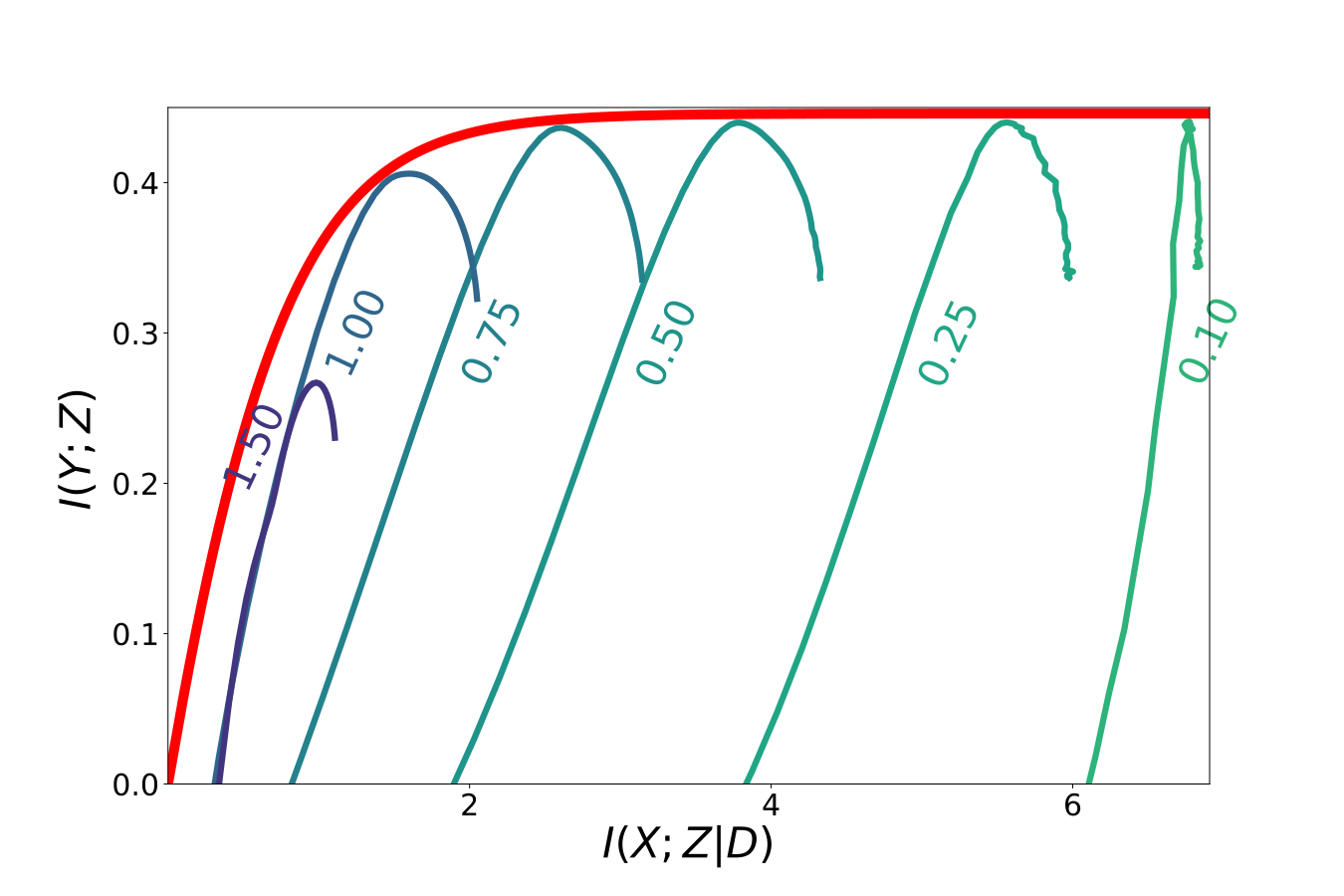} \label{fig:inf_plane_gauss_relu}}
  \subfigure[\Erf]{\includegraphics[width=0.45\textwidth]{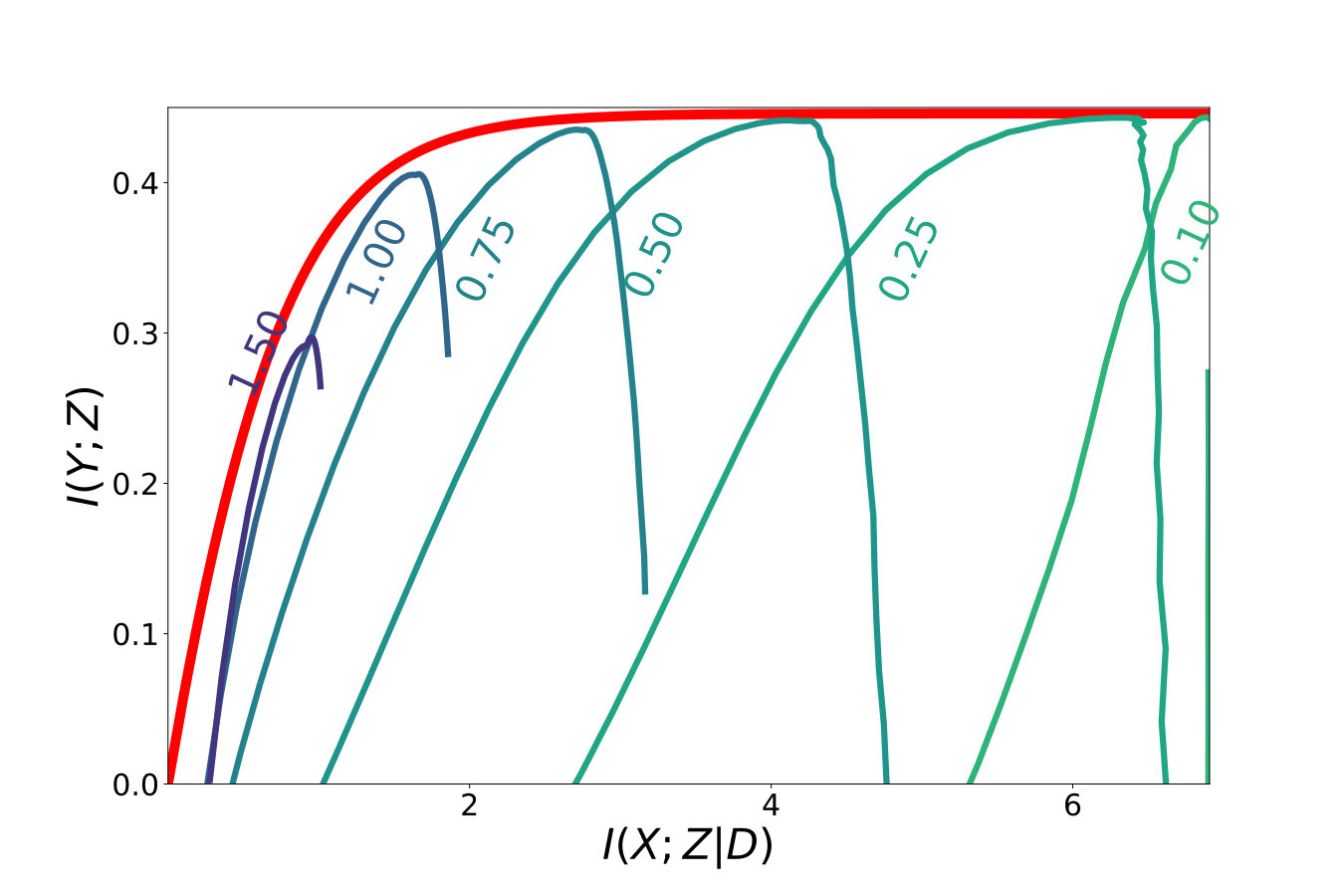}\label{fig:inf_plane_gauss_erf}}
        \caption{Trajectories of the (bounds on) MI between the representation $Z$ and the input $X$ versus time.
        Curves differ only in their initial weight variance.
        The red line is the optimal IB as predicted by theory.  
        Our estimate for $I(Z;X)$ is upper bounded by the log of the batch size ( $\log 1000 = 6.9$.)}
  \label{fig:inf_plane_gauss}
\end{figure}

In~\cref{fig:inf_plane_gauss} we show the performance of these networks on the information plane.
The $x$-axis shows a variational lower bound on the complexity of the learned representation: $I(Z;X|D)$.
The $y$-axis shows a variational lower bound on learned relevant information: $I(Y;Z)$.
For details on the calculation of the MI estimates see~\cref{sec:info}.
The curves show trajectorites of the networks' representation
as time varies from $\tau = 10^{-2}$ to $\tau = 10^{10}$
for different weight variances (the bias variance in all networks was fixed to 0.01).
The red line is the optimal theoretical IB bound.  

There are several features worth highlighting.  First, we emphasize the somewhat surprising
result that, as time goes to infinity, the MI between an infinite ensemble
of infinitely-wide neural networks output and their input is finite and quite
small.  Even though every individual network provides a seemingly rich deterministic
representation of the input, when we marginalize over the random initialization, the ensemble compresses
the input quite strongly.
The networks overfit at late times. 
For \Erf\ networks, the more complex representations ($I(Z;X|D)$) overfit more.
With optimal early stopping, over a wide range,
these models achieve a near optimal trade-off in prediction versus compression.
Varying the initial weight variance controls the amount of information the ensemble extracts.

Next, we repeat the result of the previous section on the MNIST dataset~\citep{mnist}. 
Unlike the normal setup we turn MNIST into a binary regression task for the parity of the digit (even or odd).
The network this time is a standard two-layer convolutional neural network with $5 \times 5$ filters
and either \ReLU\ or \Erf\ activation functions.

\begin{figure}[htb]
  \centering
  \subfigure[ReLU]{\includegraphics[width=0.45\linewidth]{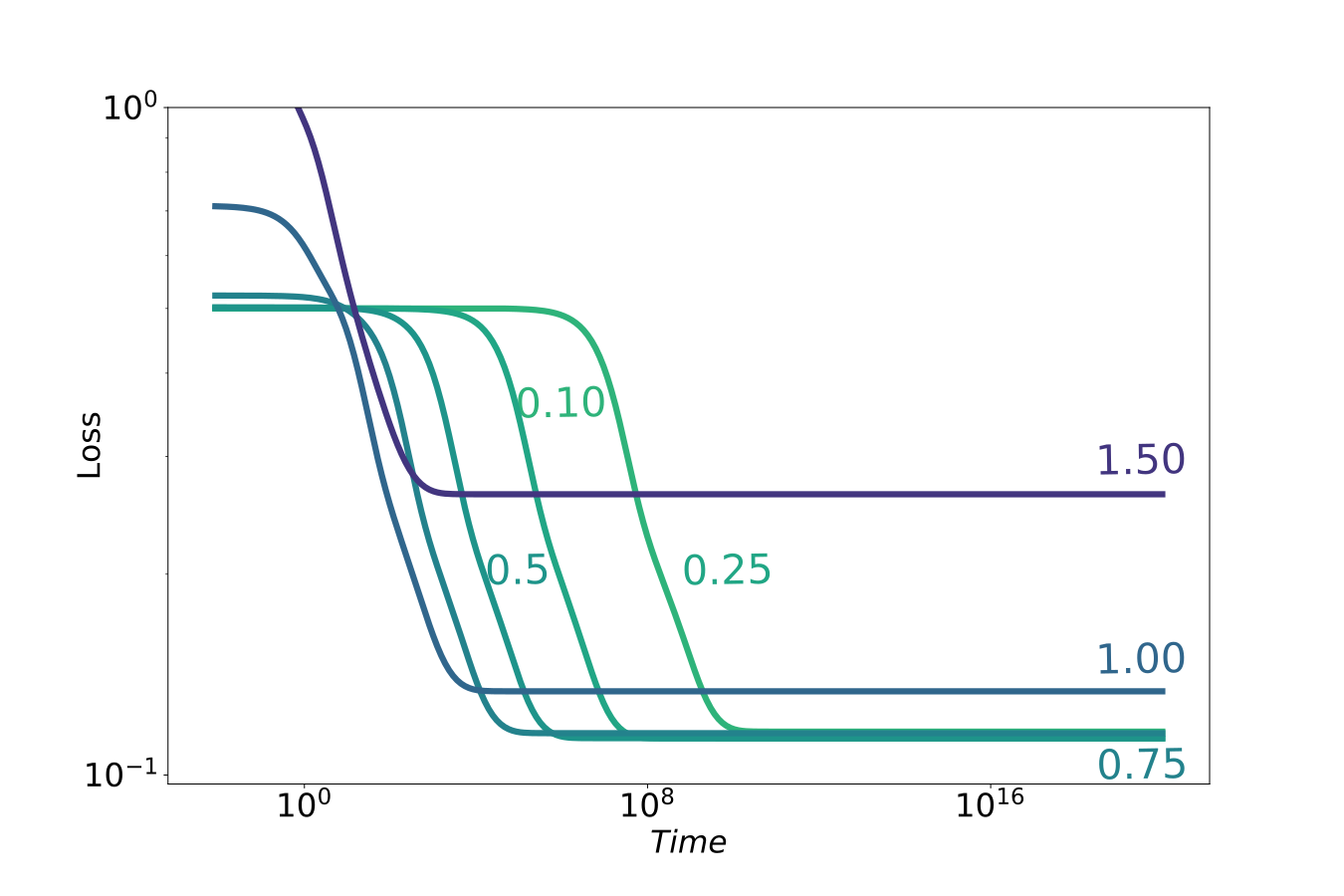} \label{fig:loss_vs_time_mnist_relu}}
  \subfigure[ERF]{\includegraphics[width=0.45\linewidth]{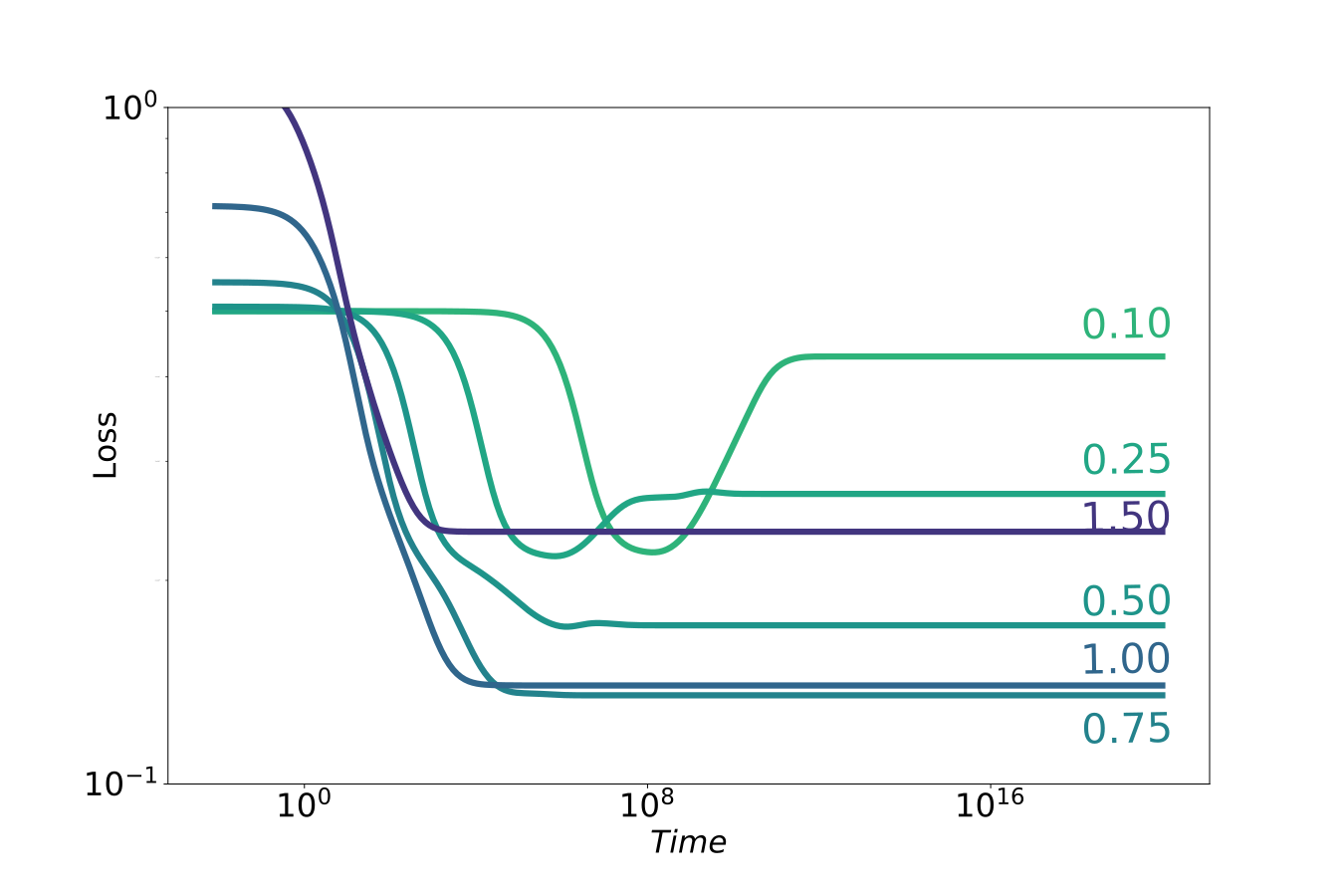}\label{fig:loss_vs_time_mnist_erf}} \\
  \subfigure[ReLU]{\includegraphics[width=0.45\textwidth]{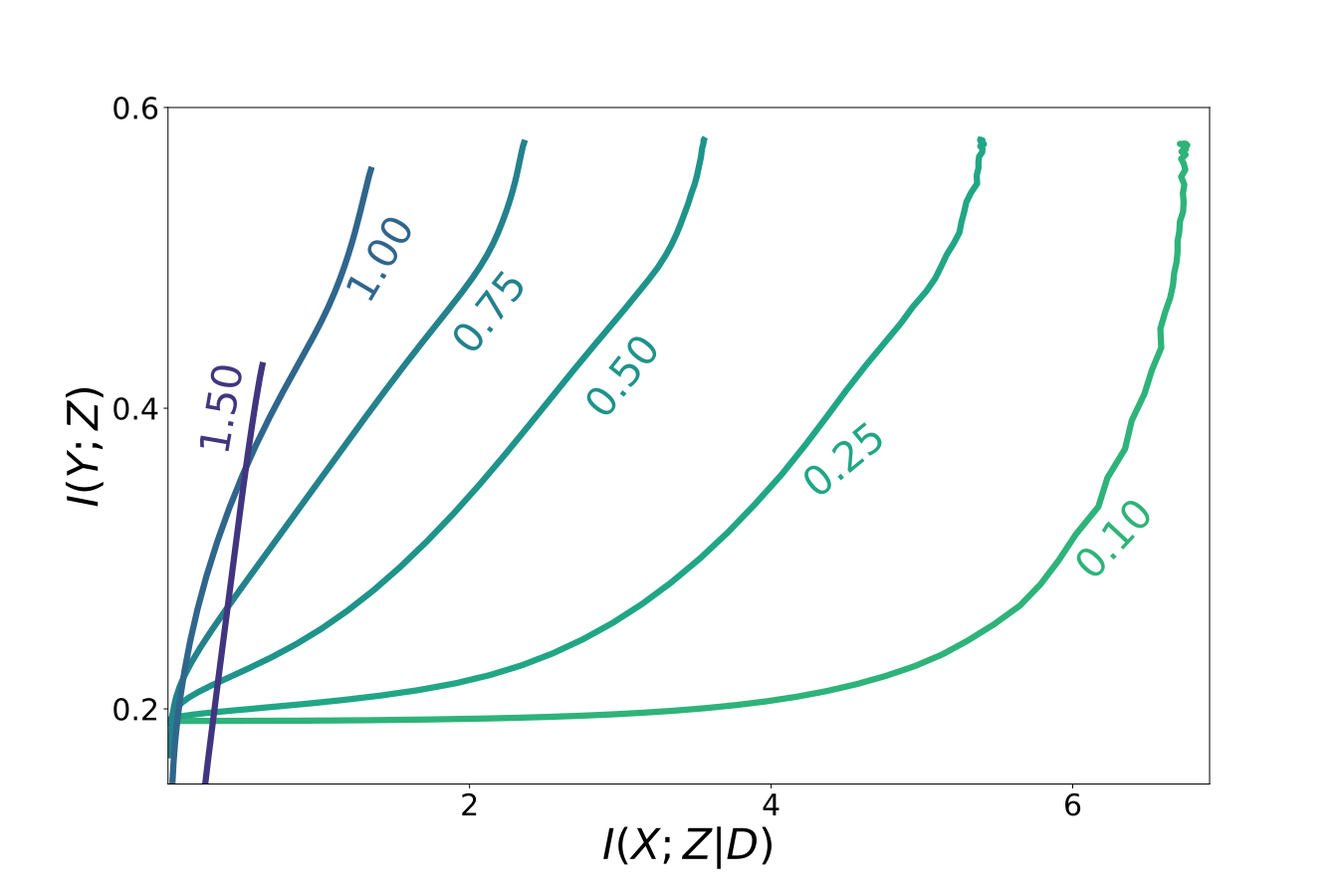} \label{fig:loss_vs_ws_mnist_relu}}
  \subfigure[ERF]{\includegraphics[width=0.45\textwidth]{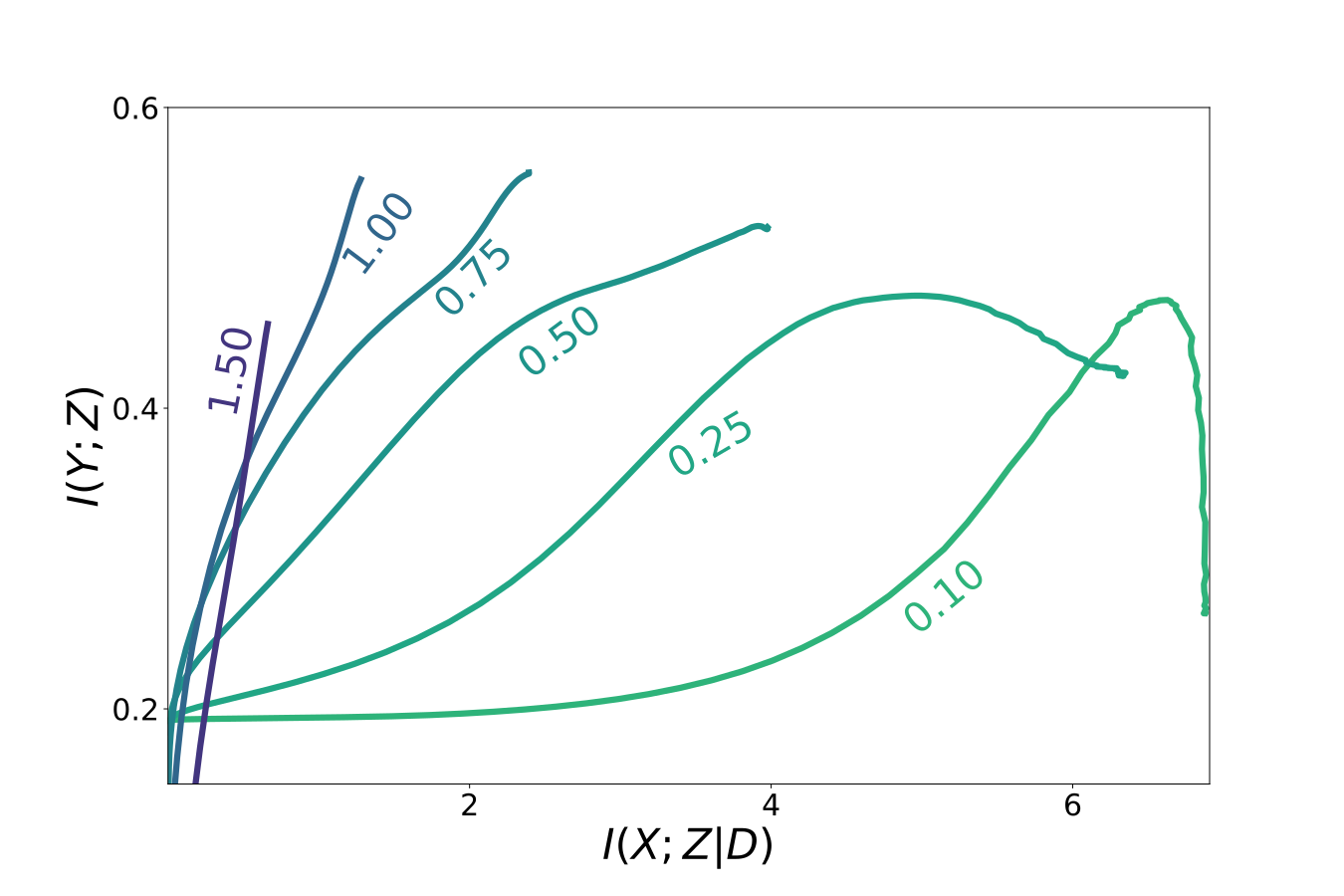}\label{fig:loss_vs_ws_mnist_erf}}
        \caption{Loss as function of time and information plan trajectories for different initial weights' variances on
        MNIST.}
         \label{fig:mnist}

\end{figure}

\Cref{fig:mnist} shows the results.
Unlike in the jointly Gaussian
dataset case, here both networks show some region of initial weight variances
that do not overfit in the sense of demonstrating any advantage from early stopping.  
The \Erf\ network at higher variances does show overfitting at low initial weight variances,
but the \ReLU\ network does not.
Notice that in the information plane, the \Erf\ network shows
overfitting at higher representational complexities ($I(Z;X)$ large),
while the \ReLU\ network does not.

\section{Conclusion}

Infinite ensembles of infinitely-wide neural networks provide an interesting model family.
Being linear in their parameters they permit a high number
of tractable calculations of information-theoretic quantities and their bounds.
Despite their simplicity, they still can achieve good generalization performance~\citep{cando}.
This challenges existing claims for the purported
connections between information theory and generalization in deep neural networks.
In this preliminary work, we laid the
ground work for a larger-scale empirical and theoretical study of generalization
in this simple model family.  
Given that real networks approach this family in their 
infinite width limit, we believe a better understanding of generalization in the NTK limit
will shed light on generalization in deep neural networks.


\bibliography{bib}

\begin{thebibliography}{20}
\providecommand{\natexlab}[1]{#1}
\providecommand{\url}[1]{\texttt{#1}}
\expandafter\ifx\csname urlstyle\endcsname\relax
  \providecommand{\doi}[1]{doi: #1}\else
  \providecommand{\doi}{doi: \begingroup \urlstyle{rm}\Url}\fi

\bibitem[{Achille} and {Soatto}(2017)]{emergence}
A.~{Achille} and S.~{Soatto}.
\newblock {Emergence of Invariance and Disentangling in Deep Representations}.
\newblock \emph{Proceedings of the ICML Workshop on Principled Approaches to
  Deep Learning}, 2017.

\bibitem[Achille and Soatto(2019)]{whereinfo}
Alessandro Achille and Stefano Soatto.
\newblock Where is the information in a deep neural network?, 2019.

\bibitem[Alemi et~al.(2016)Alemi, Fischer, Dillon, and Murphy]{vib}
Alexander~A Alemi, Ian Fischer, Joshua~V Dillon, and Kevin Murphy.
\newblock Deep variational information bottleneck.
\newblock \emph{arXiv:1612.00410}, 2016.
\newblock URL \url{http://arxiv.org/abs/1612.00410}.

\bibitem[Amjad and Geiger(2018)]{hownot}
Rana~Ali Amjad and Bernhard~C Geiger.
\newblock How (not) to train your neural network using the information
  bottleneck principle.
\newblock \emph{arXiv preprint arXiv:1802.09766}, 2018.

\bibitem[Arora et~al.(2019)Arora, Du, Li, Salakhutdinov, Wang, and Yu]{cando}
Sanjeev Arora, Simon~S. Du, Zhiyuan Li, Ruslan Salakhutdinov, Ruosong Wang, and
  Dingli Yu.
\newblock Harnessing the power of infinitely wide deep nets on small-data
  tasks, 2019.

\bibitem[Banerjee(2006)]{bayesianbounds}
Arindam Banerjee.
\newblock On bayesian bounds.
\newblock In \emph{Proceedings of the 23rd international conference on Machine
  learning}, pages 81--88. ACM, 2006.

\bibitem[Bassily et~al.(2017)Bassily, Moran, Nachum, Shafer, and
  Yehudayoff]{littlebits}
Raef Bassily, Shay Moran, Ido Nachum, Jonathan Shafer, and Amir Yehudayoff.
\newblock Learners that use little information, 2017.

\bibitem[Chechik et~al.(2005)Chechik, Globerson, Tishby, and Weiss]{gaussib}
Gal Chechik, Amir Globerson, Naftali Tishby, and Yair Weiss.
\newblock Information bottleneck for gaussian variables.
\newblock \emph{Journal of machine learning research}, 6\penalty0
  (Jan):\penalty0 165--188, 2005.

\bibitem[Chen et~al.(2018)Chen, Rubanova, Bettencourt, and Duvenaud]{neuralode}
Ricky T.~Q. Chen, Yulia Rubanova, Jesse Bettencourt, and David Duvenaud.
\newblock Neural ordinary differential equations, 2018.

\bibitem[Cover and Thomas(2012)]{coverthomas}
Thomas~M Cover and Joy~A Thomas.
\newblock \emph{Elements of information theory}.
\newblock John Wiley \& Sons, 2012.

\bibitem[Jacot et~al.(2018)Jacot, Gabriel, and Hongler]{ntk}
Arthur Jacot, Franck Gabriel, and Cl{\'e}ment Hongler.
\newblock Neural tangent kernel: Convergence and generalization in neural
  networks.
\newblock In \emph{Advances in neural information processing systems}, pages
  8571--8580, 2018.

\bibitem[Kolchinsky et~al.(2018)Kolchinsky, Tracey, and Van~Kuyk]{brendan}
Artemy Kolchinsky, Brendan~D Tracey, and Steven Van~Kuyk.
\newblock Caveats for information bottleneck in deterministic scenarios.
\newblock \emph{arXiv preprint arXiv:1808.07593}, 2018.

\bibitem[Kunstner et~al.(2019)Kunstner, Balles, and Hennig]{fisher}
Frederik Kunstner, Lukas Balles, and Philipp Hennig.
\newblock Limitations of the empirical fisher approximation, 2019.

\bibitem[LeCun and Cortes(2010)]{mnist}
Yann LeCun and Corinna Cortes.
\newblock {MNIST} handwritten digit database.
\newblock 2010.
\newblock URL \url{http://yann.lecun.com/exdb/mnist/}.

\bibitem[{Lee} et~al.(2019){Lee}, {Xiao}, {Schoenholz}, {Bahri}, {Novak},
  {Sohl-Dickstein}, and {Pennington}]{widelinear}
Jaehoon {Lee}, Lechao {Xiao}, Samuel~S. {Schoenholz}, Yasaman {Bahri}, Roman
  {Novak}, Jascha {Sohl-Dickstein}, and Jeffrey {Pennington}.
\newblock {Wide Neural Networks of Any Depth Evolve as Linear Models Under
  Gradient Descent}.
\newblock \emph{arXiv e-prints}, art. arXiv:1902.06720, Feb 2019.

\bibitem[Novak et~al.(2020)Novak, Xiao, Hron, Lee, Alemi, Sohl-Dickstein, and
  Schoenholz]{neuraltangents}
Roman Novak, Lechao Xiao, Jiri Hron, Jaehoon Lee, Alexander~A. Alemi, Jascha
  Sohl-Dickstein, and Samuel~S. Schoenholz.
\newblock Neural tangents: Fast and easy infinite neural networks in python.
\newblock In \emph{International Conference on Learning Representations}, 2020.
\newblock URL \url{https://openreview.net/forum?id=SklD9yrFPS}.

\bibitem[Poole et~al.(2019)Poole, Ozair, van~den Oord, Alemi, and
  Tucker]{vmibounds}
Ben Poole, Sherjil Ozair, A{\"{a}}ron van~den Oord, Alexander~A. Alemi, and
  George Tucker.
\newblock On variational bounds of mutual information.
\newblock \emph{CoRR}, abs/1905.06922, 2019.
\newblock URL \url{http://arxiv.org/abs/1905.06922}.

\bibitem[Saxe et~al.(2018)Saxe, Bansal, Dapello, Advani, Kolchinsky, Tracey,
  and Cox]{saxe}
Andrew~Michael Saxe, Yamini Bansal, Joel Dapello, Madhu Advani, Artemy
  Kolchinsky, Brendan~Daniel Tracey, and David~Daniel Cox.
\newblock On the information bottleneck theory of deep learning.
\newblock In \emph{International Conference on Learning Representations}, 2018.
\newblock URL \url{https://openreview.net/forum?id=ry_WPG-A-}.

\bibitem[Shwartz-Ziv and Tishby(2017)]{blackbox}
Ravid Shwartz-Ziv and Naftali Tishby.
\newblock Opening the black box of deep neural networks via information.
\newblock \emph{arXiv preprint arXiv:1703.00810}, 2017.

\bibitem[Tishby and Zaslavsky(2015)]{tishbydeep}
Naftali Tishby and Noga Zaslavsky.
\newblock Deep learning and the information bottleneck principle.
\newblock In \emph{2015 IEEE Information Theory Workshop (ITW)}, pages 1--5.
  IEEE, 2015.

\end{thebibliography}

\clearpage
\appendix

\section{Background on the NTK}
\label{sec:ntk}

Infinitely-wide neural networks behave as though they were linear in their parameters~\citep{widelinear}:
\begin{equation}
    z(x, \theta) = z_0(x) + \frac{\partial z_0}{\partial \theta} (\theta - \theta_0) \quad z_0(x) \equiv z(x, \theta_0).
\end{equation}
This makes them particularly analytically tractable.  
An infinitely-wide neural network, trained by gradient flow to minimize squared loss admits
a closed form expression for evolution of its predictions as a function of time:
\begin{equation}
    z(x, \tau) =  z_0(x) - \Theta(x, \X) \Theta^{-1} \left( I - e^{-\tau \Theta} \right)(z_0(\X) - \Y) .
\end{equation}
Here $z$ denotes the output of our neural network acting on the input $x$.  $\tau$ is a
dimensionless representation of the time of our training process.  $\X$ denotes the
whole training set of examples, with their targets $\Y$.  $z_0(x) \equiv z(x,\tau=0)$ denotes the neural networks
output at initialization.
The evolution is governed by the \emph{neural tangent kernel} (NTK) $\Theta$~\citep{ntk}.
For a finite width network, the NTK corresponds to $JJ^T$, the gram matrix of neural network
gradients. 
As the width of a network increases to infinity, this kernel converges in probability
to a fixed value.
There exist tractable ways to calculate
the exact infinite-width kernel for wide classes of neural networks~\citep{neuraltangents}.
The shorthand $\Theta$ denotes the kernel function evaluated on the train data ($\Theta \equiv \Theta(\X, \X)$).

Notice that the behavior of infinitely-wide neural networks trained with gradient flow and
squared loss is just a time-dependent affine transformation of their initial predictions.
As such, if we now imagine forming an infinite ensemble of such networks as we vary
their initial weight configurations, if those weights are sampled from a Gaussian distribution,
the law of large numbers enforces that the distribution of outputs
of the ensemble of networks at initialization is Gaussian, conditioned on its input.  Since the
evolution is an affine transformation of the initial predictions, the predictions remain Gaussian
at all times. For more details see~\citet{widelinear}.
\begin{align}
    p(z|x) &\sim \N(\mu(x,\tau), \Sigma(x,\tau)) 
    \label{eqn:repapp} \\
    \mu(x,\tau) &= \Theta(x, \X) \Theta^{-1} \left( I - e^{-\tau \Theta} \right) \Y\\
    \Sigma(x,\tau) &=  \K(x, x) + \Theta(x, \X) \Theta^{-1} \left(I - e^{-\tau \Theta} \right) \left( \K \Theta^{-1} \left( I - e^{-\tau \Theta} \right) \Theta(\X, x) - 2 \K (\X, x) \right).
\end{align}
Here, $\K$ denotes yet another kernel, the \emph{neural network gaussian process} kernel (NNGP). For a finite
width network, the NNGP corresponds to the 
expected gram matrix of the outputs: $\mathbb{E}\left[ z z^T \right]$. In the infinite width limit, this concentrates
on a fixed value.
Just as for the NTK, the NNGP can be tractably computed~\citep{neuraltangents}, and should be
considered just a function of the neural network architecture.

\section{Gaussian Dataset}
\label{sec:gaussian}

For our experiments we used a jointly Gaussian dataset, for which there is an analytic solution
for the optimal representation~\citep{gaussib}.

Imagine a jointly Gaussian dataset, where we have $x_{ij} = L^x_{jk} \epsilon^x_{ik}$ with $\epsilon \sim \mathcal{N}(0, 1)$.  Make $y$ just an affine projection of $x$ with added noise.  
\begin{equation}
    y_{ij} = L^y_{jk}\epsilon_{ik} + A_{jk} x_{ik} = L^y_{jk}\epsilon^y_{ik} + A_{jk} L^x_{km}\epsilon_{im}^x.
\end{equation}
Both $x$ and $y$ will be mean zero.  We can compute their covariances.
\begin{equation} 
\Sigma^x_{jk} = \langle x_{ij} x_{ik} \rangle = \langle L_{jm}\epsilon_{im} L_{kl}\epsilon_{il} \rangle = L_{jm}L_{kl} \delta_{ml} = L_{jm}L_{km}
\end{equation}
Next look at the covariance of $y$.
\begin{align*}
    \Sigma^y_{jk} &= \langle y_{ij} y_{ik} \rangle \\ 
    &=  \left\langle \left(L^y_{jl}\epsilon^y_{il} + A_{jl} L^x_{lm}\epsilon_{im}^x\right)
    \left( L^y_{kn}\epsilon^y_{in} + A_{kn} L^x_{no}\epsilon_{io}^x\right) \right\rangle \\
 &=  L^y_{jl} L^y_{kn} \delta_{ln} + A_{jl}L^x_{lm}A_{kn}L_{no}\delta_{mo} \\
 &= L^y_{jn}L^y_{kn} + A_{jl}\Sigma^x_{ln} A_{kn}
\end{align*}

For the cross covariance:
\begin{align*}
    \Sigma^{xy}_{jk} &= \left\langle x_{ij} y_{ik} \right\rangle \\
    &=  \left\langle L^x_{jm} \epsilon^x_{im}
    \left( L^y_{kn}\epsilon^y_{in} + A_{kn} L^x_{no}\epsilon_{io}^x\right) \right\rangle \\
    &= L^x_{jm}A_{kn}L^x_{no} \delta_{mo} \\
    &= L^x_{jm}A_{kn} L^x_{nm} = \Sigma^x_{jn} A_{kn}
\end{align*}

So we have for our entropy of $x$:
\[ 
    H(X) = \frac {n_x}2 \log (2 \pi e) + n_x \log \sigma_x
\]
\[ 
    H(Y|X) = \frac{n_y}2 \log (2\pi e) + n_y \log \sigma_y 
\]

as for the marginal entropy, we will assume the SVD decomposition $A = U \Sigma V^T$
\[
    H(Y) = \frac{n_y}{2} \log (2\pi e) + \frac{n_y}{2} \log  \left| \sigma_y^2 I + \sigma_x^2 A A^T \right|=
    \frac{n_y}{2} \log (2\pi e) + \frac{1}{2} \sum_i \log \left( \sigma_y^2 + \sigma_x^2 \Sigma_i^2 \right)
\]

So, solving for the mutual information between $x$ and $y$ we obtain:

\[
    I(X;Y) = H(Y) - H(Y|X) = \frac 12 \sum_{i} \log \left( 1 + \frac{\sigma_x^2 \Sigma_i^2}{\sigma_y^2} \right)
\]

\section{Information Metrics}
\label{sec:info}

Having a tractable form for the representation of the ensemble of infinitely-wide networks enables us to compute several information-theoretic quantities of interest.
This already sheds some light on previous attempts to explain generalization in neural networks,
and gives us candidates for an empirical investigation into
quantities that can predict generalization.

\subsection{Loss}
\label{sec:loss}

In order to compute the expected loss of our ensemble, we need to marginalize out the stochasticity in the output
of the network.  Training with squared loss is equivalent
to assuming a Gaussian observation model $p(y|z) \sim \N(0, 1)$. 
We can marginalize out our representation to obtain
\begin{equation}
    q(y|x) = \int dz\, q(y|z)p(z|x) \sim \N( \mu(x, \tau), I + \Sigma(x, \tau) ).
\end{equation}
This enables us to compute the marginal predictive loss in closed form, if desired.  Regrettably, we did not look at that quantity here.  For this work we focused on its upper bound, the expected log loss over samples of $Z$.
The expected log loss has contributions both from the square loss of the mean prediction, as well
as a term which couples to the trace of the covariance:
\begin{equation}
    \mathbb{E}\left[ \log q(y|z) \right] = \frac 12 \mathbb{E}\left[ ( y - z(x, \tau ))^2  \right]= \frac 12 ( y - \mu( x, \tau))^2 + \frac 12 \Tr \Sigma(x,\tau) - \frac k 2 \log 2 \pi
\end{equation}
here $k$ is the dimensionality of $y$.

\subsection{\texorpdfstring{$I(Z;Y)$}{I(Z;Y)}}
\label{sec:izy}

While the MI between the network's output and the targets is intractable in
general, we can obtain a tractable variational lower bound:~\citep{vmibounds}
\begin{equation}
    I(Z; Y) = \mathbb{E}\left[ \log \frac{p(y|z)}{p(y)} \right] 
    \leq \mathbb{E}\left[ \log \frac{q(y|z)}{p(y)} \right] = H(Y) + \mathbb{E}\left[ \log q(y|z) \right]
\end{equation}

\subsection{\texorpdfstring{$I(Z;X|D)$}{I(Z;X|D)}}
\label{sec:izx}

The MI between the input ($X$)
and output ($Z$) of our network,
conditioned on the dataset ($D$) is:
\begin{equation}
    I(Z;X | D) = \mathbb{E}\left[ \log \frac{p(z|x,D)}{p(z|D)} \right].
\end{equation}
This requires knowledge of the marginal distribution $p(z|D)$.  Without knowledge of 
$p(x)$, this is in general intractable, but
there exist simple tractable multi sample 
upper and lower bounds~\citep{vmibounds}:
\begin{equation}
    \frac 1 N \sum_i \log \frac{p(z_i|x_i, D)}{\frac 1 {N} \sum_{j} p(z_i| x_j, D)} \leq
    I(Z;X | D) 
    \leq \frac 1 N \sum_i \log \frac{p(z_i|x_i, D)}{\frac 1 {N-1} \sum_{j \neq i} p(z_i| x_j, D)}.
\end{equation}
In this work, we show the minibatch lower bound
estimates, which are upper bounded themselves by
the log of the batch size.

\subsection{\texorpdfstring{$I(Z;D|X)$}{I(Z;D|X)}}
\label{sec:izd}

We can also estimate a variational upper bound on the
MI between the representation 
of our networks and the training dataset.
\begin{equation}
    I(Z;D|X) = \mathbb{E}\left[ \log \frac{p(z|x,D)}{p(z|x)} \right] \leq \mathbb{E}\left[ \log \frac{p(z|x,D)}{p_0(z|x)} \right].
\end{equation}
Here, the MI we extract from the dataset 
involves the expected log ratio of our posterior distribution
of outputs to the marginal over all possible datasets.
Not knowing the data distribution, this is intractable in
general, but we can variationally upper bound it with an
approximate marginal.  A natural candidate is the prior
distribution of outputs, for which we have a tractable estimate.

\subsection{Fisher}
\label{sec:fisher}

Infinitely-wide networks behave as though they were linear in their parameters with a fixed Jacobian.  This leads to a trivially flat information geometry.  For squared loss
the true Fisher can be computed simply as $F = J^T J$~\citep{fisher}.  
While the trace  of the Fisher information has recently been proposed as an important quantity
for controlling generalization in neural networks~\citep{whereinfo}, for infinitely
wide networks we can see that the trace of the fisher is the same as the trace of the
NTK, which is a constant and does not evolve with time ($\Tr F = \Tr J^T J =\Tr J J^T = \Tr \Theta$).
In so much as infinite ensembles of infinitely-wide neural networks generalize, the degree to which
they do or do not cannot be explained by the time evolution of the trace of the
Fisher given that the trace of the Fisher does not evolve.

\subsection{Parameter Distance}
\label{sec:dist}

How much do the parameters of an infinitely-wide network change?  
Other work~\citep{widelinear}
emphasizes that the relative Frobenius norm change of the 
parameters over the course of training vanishes in the limit of infinite width.
This is in fact a justification for the linearization becoming more
accurate as the network becomes wider. 
But is it thus fair to say the parameters are not
changing?  Instead of looking at the Frobenius norm we can investigate the
\emph{length} of the parameters path over the course of training.  This 
reparameterization independent notion of distance utilizes the 
information geometric metric provided by
the Fisher information:
\begin{equation}
    L(\tau) = \int_0^\tau ds =
    \int_0^\tau d\tau\, \sqrt{\dot\theta_\alpha(\tau) g_{\alpha \beta} \dot\theta_\beta(\tau) }  
    =  \int_0^\tau d\tau\, \left\lVert \Theta  e^{-\tau \Theta} (z_0(\X) -\Y) \right\rVert
\end{equation}
The length of the trajectory in parameter space is the integral of a norm of our
residual at initialization projected along $\Theta e^{-\tau \Theta}$.  This integral is both 
positive and finite even as $t \to \infty$.  To get additional understanding into the
structure of this term, we can consider its expectation over the ensemble, where
we can use Jensen's inequality to bound the expectation of trajectory lengths.  Since we
know that at initialization $z_0(\X) \sim \N(0, \K)$ we obtain further simplifications:
\begin{align}
    \mathbb{E}[L(\tau)]^2 &\leq \mathbb{E}[L^2(\tau)] =
    \int_0^\tau d\tau\, \mathbb{E}\left[ (z_0(\X) - \Y)^T \Theta^2 e^{-2\tau \Theta} (z_0(\X) - \Y) \right]  \\
     &=  \frac 12 \mathbb{E}\left[ (z_0(\X) - \Y)^T \Theta \left( 1 - e^{-2\tau \Theta} \right) (z_0(\X) - \Y) \right]  \\
     &= \frac{1}{2} \left[ \Tr{\left( \K \Theta \left( 1 - e^{-2\tau \Theta}\right)\right)} + \Y^T\Theta \left(1 - e^{-2\tau \Theta}\right) \Y \right].
\end{align}

\subsection{\texorpdfstring{$\kl{p(\theta|D)}{p_0(\theta)}$}{KL[p(θ|D), p_0(θ)]}}
\label{sec:itd}

The MI between the parameters and the dataset $I(\theta; D)$ has been shown to control for overfitting~\citep{littlebits}. We can generate a variational upperbound on this 
quantity by consider the KL divergence between the posterior distribution of our parameters
and the prior distribution $\kl{p(\theta|D)}{p_0(\theta)}$ a quantity that itself has been
shown to provide generalization bounds in PAC Bayes frameworks~\citep{emergence}. For our networks, 
the prior distribution is known and simple, but the posterior distribution can be quite rich. 
However, we can use the \emph{instantaneous change of variables} formula~\citep{neuralode}
\begin{equation}
   \log p(\theta_\tau) = \log p(\theta_0) - \int_0^\tau d\tau\, \Tr \left( \frac{\partial \dot \theta}{\partial \theta} \right),
\end{equation}
which gives us a value for the log likelihood the parameters of a trained model at any point in time
in terms of its initial log likelihood and the integral of the trace of the kernel governing its
time evolution.  For our infinitely-wide neural networks this is tractable:
%
%
\begin{align}
    I(\theta; D) &\leq 
    \mathbb{E}_{p(\theta_\tau)}\left[ \log p(\theta_\tau) - \log p_0(\theta_\tau)  \right]\\
    &=  \mathbb{E}_{p(\theta_0)}\left[ \log p(\theta_0) - \log p_0(\theta_\tau) - \int_0^\tau d\tau\, \Tr\left( \frac{\partial \dot\theta}{\partial \theta}\right) \right]\\
    &=  \mathbb{E}_{p(\theta_0)}\left[ \log p(\theta_0) - \log p_0(\theta_0 + \Delta\theta_\tau) + \tau \Tr \Theta \right]\\
    &=  \mathbb{E}_{p(\theta_0)}\left[ -\frac 12 \theta_0^2 + \frac 12 \left( \theta_0 + \Delta\theta_\tau \right)^2 \right] + \tau \Tr \Theta \\
    &=  \mathbb{E}_{p(\theta_0)}\left[  \theta_0^T \Delta\theta_\tau + (\Delta\theta_\tau   )^2 \right] + \tau \Tr \Theta \\
    &=  \mathbb{E}_{p(\theta_0)}\left[  \theta_\tau^T \Delta\theta_\tau \right] + \tau \Tr \Theta \\
    &=  \mathbb{E}_{p(\theta_0)}\left[  (z_0(\X) - \Y)^T (I-e^{-\tau\Theta})^2\Theta^{-1}(z_0(\X)-\Y) \right] + \tau \Tr \Theta \\
    &=  \Tr \left(\K \Theta^{-1} (I-e^{-\tau \Theta})^2 \right) + \Y^T \Theta^{-1} (I-e^{-\tau \Theta})^2 \Y  + \tau \Tr \Theta .
\end{align}
This tends to infinity as the time goes to infinity.  This renders the usual PAC-Bayes style
generalization bounds trivially vacuous for the generalization of infinitely wide neural networks
at late times.  Yet, infinite networks can generalize well~\citep{cando}.

\subsection{\texorpdfstring{$I(\theta; D)$}{I(θ; D)}}
\label{sec:itdreal}

The real thing.

Consider whether we can get a lower bound too.

\section{Additional Empirical Results}
 \label{sec:more_results}
 
\begin{figure}[htb]
   \centering
   \subfigure[ReLU]{\includegraphics[width=0.45\textwidth]{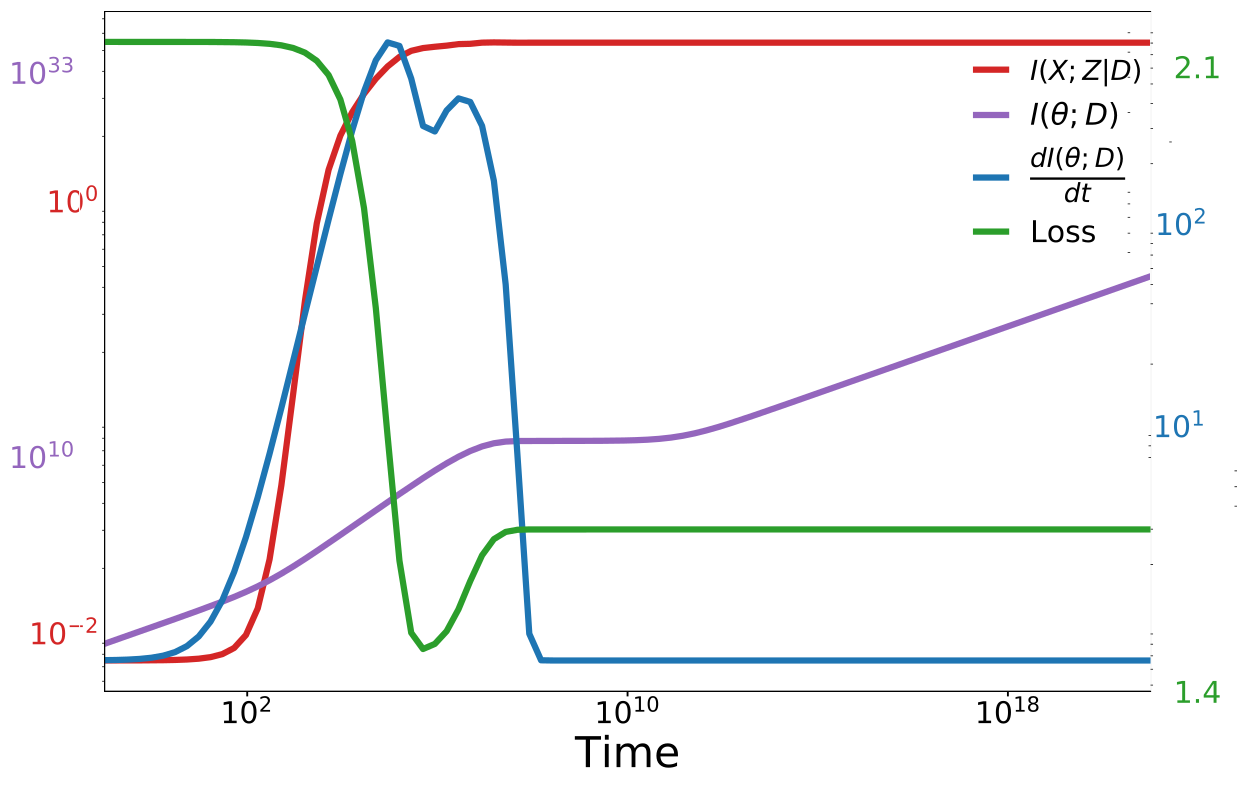} \label{fig:i_theta_data_gaus_relu}}
   \subfigure[Erf]{\includegraphics[width=0.45\textwidth]{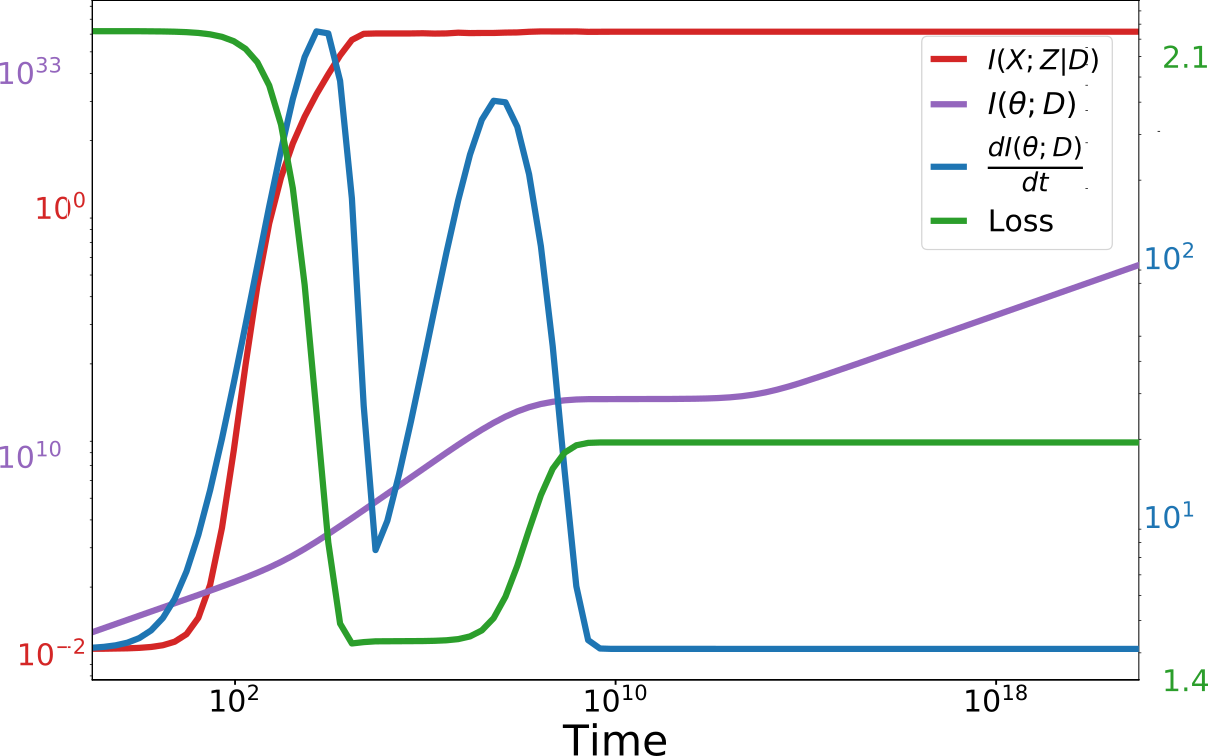}\label{ig:i_theta_data_gaus_erf}}

        \caption{$I(X;Z|D)$, $I(\theta;D)$, $\frac{dI(\theta;D)}{dt}$ and the loss as function of time on our Gaussian dataset.}
         \label{fig:i_theta_data_gaus}
 \end{figure}

\Cref{fig:i_theta_data_gaus} shows
$I(X;Z|D)$, $I(\theta;D)$, $\frac{dI(\theta;D)}{dt}$ and the loss as function of time for a fixed initial weight's variance ($\sigma_w = 0.25$).  (in log-log scale, notice that $y$-axes
are different for each measure). For both the \ReLU\ and \Erf\ networks,
we see clear features in each plot near the optimal test loss.
 
\end{document}